\documentclass[letterpaper,twoside]{IEEEtran}
\usepackage[utf8]{inputenc}
\usepackage[T1]{fontenc}
\usepackage[cmex10]{amsmath}
\usepackage{amssymb}
\usepackage{amsfonts}
\usepackage{amsthm}
\usepackage{algorithm}
\usepackage{algorithmicx}
\usepackage{algpseudocode}
\usepackage{graphicx}
\usepackage{epsfig}
\usepackage{gensymb}
\usepackage{hyperref}
\hypersetup{
    colorlinks=true,
    linkcolor=blue,
    filecolor=magenta,      
    urlcolor=cyan,
}
\usepackage{cite}
\usepackage{tensor}
\usepackage{colortbl}
\usepackage[caption=false,font=footnotesize]{subfig}
\usepackage{color}
\input{bold_letter.sty}
\graphicspath{{figure/}}
\usepackage{bm}
\usepackage{url}
\usepackage{pgfgantt}
\usepackage{dirtytalk}
\usepackage{todonotes}
\usepackage{multirow}

\newcommand{\red}[1]{\textcolor{red}{#1}}
\newcommand{\cyan}[1]{\textcolor{cyan}{#1}}

\usepackage{tikz,xcolor,hyperref}
\definecolor{lime}{HTML}{A6CE39}
\DeclareRobustCommand{\orcidicon}{
	\hspace{-2.5mm}
	\begin{tikzpicture}
	\draw[lime, fill=lime] (0,0) 
	circle [radius=0.16] 
	node[white] {{\fontfamily{qag}\selectfont \tiny ID}};
	\draw[white, fill=white] (-0.0625,0.095) 
	circle [radius=0.007];
	\end{tikzpicture}
	\hspace{-3mm}
}

\foreach \x in {A, ..., Z}{%
	\expandafter\xdef\csname orcid\x\endcsname{\noexpand\href{https://orcid.org/\csname orcidauthor\x\endcsname}{\noexpand\orcidicon}}
}

\graphicspath{{figure/}}
\DeclareGraphicsExtensions{.eps}
\title{\LARGE \bf Learning Task-Parameterized Skills from Few Demonstrations}

\author{Jihong~Zhu,~Michael~Gienger,~and~Jens~Kober
\thanks{J.~Zhu is with Cognitive Robotics, 3mE, Delft University of Technology and Honda Research Institute, Europe}
\thanks{M. Gienger is with Honda Research Institute, Europe}
\thanks{J.~Kober is with Cognitive Robotics, 3mE, Delft University of Technology}

\thanks{This work was accepted at IEEE Robotics and Automation Letters. Copyright will be transferred soon, this version will then be updated with the IEEE copyright notice and full citation, with a link to the final, published paper in IEEE Xplore.}
}

\begin{document}


\markboth{ZHU \MakeLowercase{\textit{et al.}}: Learning Task-Parameterized Skills from Few Demonstrations, accepted by the IEEE RA-L}
{
}
\maketitle

\begin{abstract}
Moving away from repetitive tasks, robots nowadays demand versatile skills that adapt to different situations. Task-parameterized learning improves the generalization of motion policies by encoding relevant contextual information in the task parameters, hence enabling flexible task executions. However, training such a policy often requires collecting multiple demonstrations in different situations. To comprehensively create different situations is non-trivial thus renders the method less applicable to real-world problems. Therefore, training with fewer demonstrations/situations is desirable. This paper presents a novel concept to augment the original training dataset with synthetic data for policy improvements, thus allows learning task-parameterized skills with few demonstrations. Videos of the experiments are available at \url{https://sites.google.com/view/tp-gmm-from-few-demos/}
\end{abstract}


\IEEEpeerreviewmaketitle
\section{Introduction}\label{sec:intro} 
    In contrast to industrial robots that operate in cages and perform repetitive tasks, a next generation of robots is expected to have higher autonomy, the ability to operate in unstructured environments and to be adaptive in task executions. Learning from demonstration (LfD) is a promising step in this direction, enabling robots to acquire versatile motor skills without explicitly programming the motion, thus facilitating robot skill learning. 
    
    In LfD, robot motion policies are generated from an underlying model that is trained from demonstration data. How to use the data efficiently and produce policies that generalize well to new situations is at the core of robot LfD research \cite{ravichandar2020recent}. One prominent example, Task-Parameterized Gaussian Mixture Models (TP-GMM) improves generalization by encoding the task-relevant states into the task parameters and use them for generating motions in a new situation \cite{calinon2016tutorial}. In TP-GMM, the task parameters are reference frames that describe the spatial configurations of the situation. Perspectives from different reference frames are leveraged to produce a policy that adapts to the current situation.
    
    Multiple demonstrations in different situations need to be collected for training the TP-GMM. Hence, the collected observation data needs to comprise many different spatial configurations of the task to provide enough statistics for a meaningful model. This is often impracticable in practice, e.g., in a factory or household environment. Furthermore, demonstrating the task with changing parameters is more likely to introduce ambiguity in the demonstration. For instance, if there is an object that the robot needs to avoid during task execution, in TP-GMM, a reference frame will be assigned to the object. During demonstrations, it is not easy to ensure that the demonstrator always goes from the similar direction in the object frame for avoidance, thus bring in ambiguity and consequently compromise the policy \cite{franzese2020learning}. 
    
    The contribution of this paper is a concept for learning task-parameterized skills from few demonstrations. Instead of solely imitating the expert, it allows generation of synthetic demonstration data that aggregate the original dataset for improving the TP-GMM. The framework reduces the number of demonstrations needed for training task-parameterized skills, improves the data efficiency, and subsequently trims the possibility of ambiguous demonstrations, thus making the task-parameterized skill learning more appealing in practice. 
    
	In the next section, we review related works about LfD with a focus on task-parameterized learning. A brief description of the TP-GMM is presented in Sect.~\ref{sec:background}. In Sect.~\ref{sec:methods}, we describe our algorithm. The algorithm is then discussed and validated with simulation in Sect. \ref{sec:simulation}. Sect.~\ref{sec:result} showcases our algorithm on a robotic dressing assistance task. Finally in Sect.~\ref{sec:conclusion}, we conclude. 
	

\section{Related Works}\label{sec:related_work}

    Methods such as dynamical movement primitives (DMP) \cite{ijspeert2013dynamical}, probabilistic movement primitives (ProMP) \cite{paraschos2013probabilistic}, Gaussian mixture models (GMMs) \cite{muhlig2009task} and more recently conditional neural movement primitives \cite{seker2019conditional} have been used for encoding movements with LfD. These methods are known for data efficiency and can generate robot motion policy from a small number of demonstrations. Alternatively, instead of relying on expert demonstrations, reinforcement learning (RL) generates data by random exploration and finds an optimal policy by reward optimization. This is usually less data efficient than LfD-based approaches. Nevertheless, combining LfD and RL are reported to further boost data efficiency \cite{kober2014policy, nair2018overcoming}. LfD benefits from demonstration data for skill learning, nevertheless, it is also limited by the reliance on data \cite{ravichandar2020recent}. Improvement can be made consider data efficiency or policy generalization. 
    
    \textbf{For improving data efficiency in LfD}: meta-imitation learning reuses data collected for different tasks \cite{huang2019neural} thus boosts data efficiency. Another common approach for improving efficiency is synthetic data augmentation. The authors of \cite{zhao2021augmenting} show that policy improvement can also be made from a single demonstration that is augmented with noise. In robotics LfD, DART is introduced for robust imitation learning \cite{laskey2017dart}. Recently, the authors of \cite{brown2020better} present a method that adds different level of noise into the demonstration data for improving the reward learning in an inverse reinforcement learning setting.
    
    \textbf{For improving generalization in LfD}: \cite{dragan2015movement} improves original DMP by selecting a Hilbert norm that reduces the deformation in the retrieved trajectory. In a more recent research, \cite{zhou2019learning} combines DMP and ProMP and proposes Viapoints Movement Primitives that outperforms ProMP in extrapolation.  
    Task parameterized models are alternative approaches for better generalization of the policy. In these approaches, the contextual information about the task are described by task parameters, and movement is encoded with either GMMs or hidden Markov models (HMMs) \cite{calinon2016tutorial, pervez2018learning, huang2018generalized}. By combining the movement model with task parameters, TP-GMM(HMM) can produce a policy that adapts to different situations. 

    \textbf{Improvements on TP-GMM}: Multiple improvements on the original TP-GMM have been made in previous research. By learning the forcing term in DMP with TP-GMM, \cite{pervez2018learning} presents an approach that resolves the divergence problem usually associated with GMM type models. \cite{huang2018generalized} introduces weighting to TP-GMM and \cite{sena2019improving} build on the idea and proposes to infer weights from co-variances in the normal distribution. While the focus of above-mentioned improvements is on better policy generalization instead of data efficiency, our paper alternatively addresses the latter which is an equally important problem in LfD.

\section{Preliminaries}\label{sec:background}
    Below we briefly present the TP-GMM algorithm, and for an in-depth tutorial on the subject, the reader can refer to \cite{calinon2016tutorial}. We split TP-GMM into two phases: \emph{Model training} and \emph{Fusion for new situations}, where the former describes how to obtain a TP-GMM from demonstrations, and the latter applies the TP-GMM to new situations.
    
    \subsection{Model Training:}\label{sec:model_training}
    In TP-GMM, the situation states for the task is described using $N$ reference frames $\big\{\vA_n, \vb_n\big\}^N_{n = 1}$
    with $\vA_n \in SO(p)$, $\vb_n \in \mathbb{R}^p$ represents the orientation and displacement in the $n$\textsuperscript{th} reference frame of $p$ dimensions\footnote{In robotic, depending on the task, $p = 2~\text{or}~3$.}.
    
    The demonstrated motion is denoted as $\vxi$ and composed of input and output data:
     \begin{equation}\label{eq:demo_data_structure}
      \vxi = \begin{bmatrix}
      \vxi^\mathcal{I}, \vxi^\mathcal{O}
      \end{bmatrix}^T.
    \end{equation}
    For a time-based TP-GMM, the input is one dimensional time, and the outputs are the positions, while a trajectory-based TP-GMM has positions as inputs and velocities (and maybe also accelerations) as outputs.    
    
    Given $M$ demonstrated trajectories $\{\vxi_m\}^M_{m = 1}$, for the $m$\textsuperscript{th} demonstration, we assign the corresponding $N$ reference frames: $\{\vA_{m,~n}, \vb_{m,~n}\}^N_{n = 1}$ to the demonstration. Using these reference frames, we can transform each demonstration into $N$ frames. Once done, we have a dataset consisting of demonstrated trajectories seen from $N$ frames. 
    
    A TP-GMM with $K$ components can be trained from this dataset and is defined by:
    \begin{equation}\label{eq:tp_gmm_model}
        \big\{ \pi_k, \{ \vmu_k^{n},~ \vSigma_k^{n} \}^N_{n = 1} \big\}^K_{k = 1}, 
    \end{equation}
    where $\pi_k$ is the $k$\textsuperscript{th} mixing coefficient, and $\vmu_k^{n},~ \vSigma_k^{n}$ are respectively the center and covariance matrix of the $k$\textsuperscript{th} Gaussian component in frame $n$.  
    
    \subsection{Fusion for New Situations}\label{sec:fusion_for_new_sit}
    For a new situation defined by a set of $N$ references frames
    \begin{equation}\label{eq:new_frames}
        \{\hat{\vA}_n, \hat{\vb}_n\}^N_{n = 1},
    \end{equation}
    a GMM that produces the motion in the new situation can be derived using  \eqref{eq:tp_gmm_model} and \eqref{eq:new_frames} in two steps. 
    
    Step $1$ - for $n$\textsuperscript{th} new reference frame, we transform the Gaussian distributions in $n$\textsuperscript{th} frame in \eqref{eq:tp_gmm_model} into the new frame using \eqref{eq:new_frames}: 
    \begin{align}\label{eq:frame_transform}
        \begin{split}
            \hat{\vmu}^n_k = \hat{\vA}_n \vmu_k^{n} + \hat{\vb}_n,~
            \hat{\vSigma}_k^n = \hat{\vA}_n \vSigma_k^{n} \hat{\vA}^T_n
        \end{split}
    \end{align}
    We apply \eqref{eq:frame_transform} to all $K$ components in \eqref{eq:tp_gmm_model}.
    
    Step $2$ - for every component, we fuse the transformed Gaussian distributions in $N$ frames (which we obtained from \eqref{eq:frame_transform} in step $1$) by:
    \begin{align}\label{eq:general_fusion}
        \begin{split}
            {\hat{\vSigma}_k}^{-1} = \sum^N_{n=1} \hat{\vSigma}_k^{n ^ {-1}}, 
        ~\hat{\vmu}_k & = \hat{\vSigma}_k \sum^N_{n=1} \hat{\vSigma}_k^{n ^ {-1}} \hat{\vmu}^n_k.
        \end{split}
    \end{align}
  The new GMM that adapts to the new frames is then $\{ \pi_k, \hat{\vmu}_k, \hat{\vSigma}_k \}^K_{k = 1}$. Likewise in \eqref{eq:demo_data_structure}, the resulting GMM can be decomposed as:
  \begin{equation}
      \hat{\vmu}_k = \begin{bmatrix}
      \hat{\vmu}_k^\mathcal{I} \\
      \hat{\vmu}_k^\mathcal{O}
      \end{bmatrix},~
      \hat{\vSigma}_k = \begin{bmatrix}
      \hat{\vSigma}_k^\mathcal{I} & \hat{\vSigma}_k^{\mathcal{I}\mathcal{O}} \\
      \hat{\vSigma}_k^{\mathcal{O}\mathcal{I}} & \hat{\vSigma}_k^\mathcal{O}
      \end{bmatrix}.
  \end{equation}
  Given input data $\vy^\mathcal{I}$ (current time instance for time-based TP-GMM or positions for trajectory-based TP-GMM), the motion generation in the new situation is solved using Gaussian Mixture Regression (GMR) \cite{calinon2016tutorial}.


\section{Methods}\label{sec:methods}
    We present the proposed method comprehensively in this section. We start by giving an intuitive description in Sect. \ref{sec:design_considerations} then an in-depth explanation on each step of the algorithm in Sect. \ref{sec:method_details}.
    
    \subsection{Design Considerations}\label{sec:design_considerations}
\begin{figure}
        \centering
        \centering
        \subfloat[Example of a GMM in one situation]{\includegraphics[width=0.32\columnwidth]{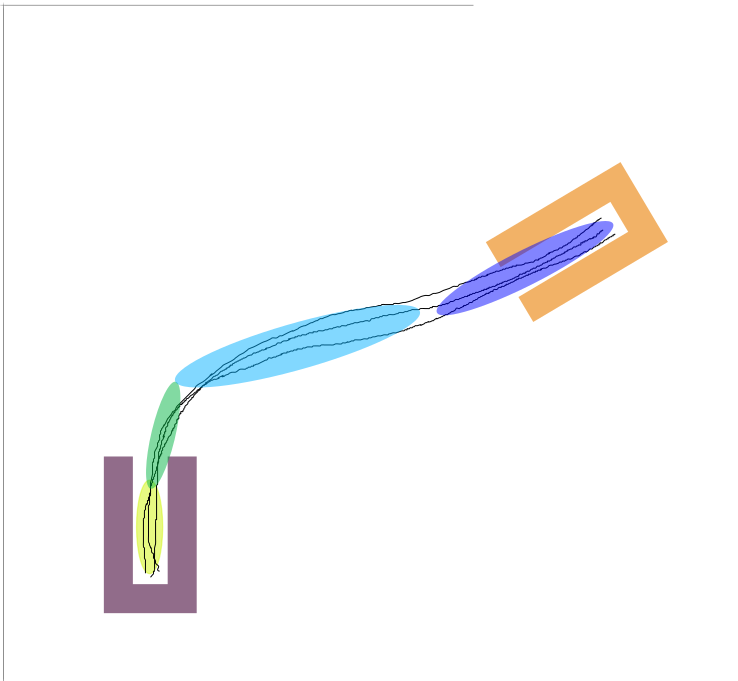}\label{fig:GMM}}
        \hfill
        \subfloat[Example of a TP-GMM with 2 reference frames and 5 situations]{\includegraphics[width=0.64\columnwidth]{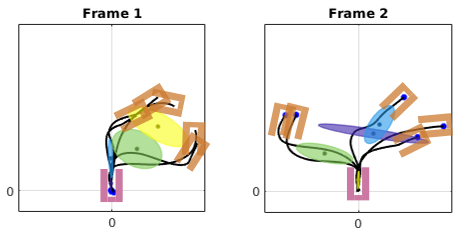}\label{fig:TP_GMM}}
        \caption{A comparison between GMM and TP-GMM for a 2D movement task: from the start (the grey U-shape box) to the goal (the yellow U-shape box)}
        \label{fig:Comparison_GMM_TPGMM}
    \end{figure}

    Rather than explicitly minimizing the difference between learned and demonstrated policy, TP-GMM relies on a good representation of data distributions in each reference frame in order to obtain a good policy in different situations. Fig. \ref{fig:Comparison_GMM_TPGMM} presents the comparison between a GMM and TP-GMM. While the former explicitly maximize the likelihood of motion data in one situation, the latter learns the data representation in each frames under different situations.
    
    
    In the model training step presented in Sect.~\ref{sec:model_training}, the learned TP-GMM in the form of \eqref{eq:tp_gmm_model} maximizes the joint likelihood of data distributions in different reference frames. Since we start with only few demonstrations, the data will be sparse, and the learned model will not be able to capture the distributions well in each frame. Subsequently, the model would fail to obtain a decent fusion for policy generation.
    
    By explicitly taking the reproduction error as the selection criteria, our algorithm generates and selects synthetic data that augments the original training data for policy improvement, thus allows learning task-parameterized skills from few demonstrations.
    
    

    
    \subsection{Learning Task Parameterized Skills from Few Demonstrations}\label{sec:method_details}
    We design Alg. \ref{alg} for learning task-parameterized skill with few demonstrations. We elaborate on the steps marked with \textbf{\emph{bold}} -- \emph{\textbf{italic}} in this section.
    \begin{algorithm}[t]
     \caption{Learning Task-Parameterized Skills from Few Demonstrations}\label{alg}
     \hspace*{\algorithmicindent} \emph{\textbf{Inputs:}} Initial $\mu$ demonstrations, $\mathcal{D}_{\text{init}}$\\
     \hspace*{\algorithmicindent} \hspace{11mm} Maximum number of demonstrations, $M$ \\
     \hspace*{\algorithmicindent} \hspace{11mm} Maximum number of iteration, $L$ \\
     \hspace*{\algorithmicindent} \emph{\textbf{Outputs:}} Final TP-GMM $\mathcal{P}_\text{final}$
        \begin{algorithmic}[1]
            \State Train a TP-GMM $\mathcal{P}$ based on $\mathcal{D}_{\text{init}}$
            \State \emph{\textbf{Cost Computation:}} $\mathcal{J}(\mathcal{P}, \mathcal{D}_{\text{init}})$ 
            \State $\text{iter} = 0,~n_{\text{d}} = \mu,~ \mathcal{D} = \mathcal{D}_{\text{init}}$
            \While{$n_{\text{d}} \leq M$ or $\text{iter} < L$}  
                \State \emph{\textbf{Generate Synthetic Data}} $\mathcal{D}_n$
                \State Aggregrate datasets: $\mathcal{D}^\prime \leftarrow \mathcal{D} + \mathcal{D}_n$
                \State Retrain the TP-GMM $\mathcal{P}^\prime$ from $\mathcal{D}^\prime$
                \State Compute the cost with new TP-GMM $\mathcal{J}^\prime(\mathcal{P}^\prime, \mathcal{D}_{\text{init}})$ 
                \If{$\mathcal{J}^\prime < \mathcal{J}$}
                    \State $\mathcal{D} = \mathcal{D}^\prime$
                    \State $\mathcal{P} = \mathcal{P}^\prime$
                    \State $n_{\text{d}} = n_{\text{d}} + 1$
                \EndIf
                \State $\text{iter} = \text{iter} + 1$
            \EndWhile
            \State $\mathcal{P}_\text{final} = \mathcal{P}$
            \State \Return $\mathcal{P}_\text{final}$
        \end{algorithmic}
     \end{algorithm}
    
    \textbf{\emph{Inputs:}} In this step, we mainly discuss the demonstrations collection. We collect $\mu$ demonstrations (where $\mu \geq 2$) $\mathcal{D}_{\text{init}}$ for the initial training of the TP-GMM. The number $\mu$ depends on the complexity of the task (i.e., tasks with more frames and less constraints may require a larger $\mu$). In order to avoid that the final policy over-fits to some local configurations and improve generalization, it is better if the demonstrations are collected in distinctive situations (see Sect. \ref{sec:result} for some discussions). Since in TP-GMM, the situations are described with reference frames, the distinctive measure is the distances between the corresponding reference frames, and difference in Euler angles that represents difference in orientations. 
    Training a TP-GMM from initial demonstrations is a standard procedure described in Sect.~\ref{sec:background}.
    
    \textbf{\emph{Cost Computation:}}  The cost is defined as the reproduction error between the TP-GMM produced policy and the expert demonstration. Therefore it takes the initial demonstration $\mathcal{D}_{\text{init}}$ and the TP-GMM $\mathcal{P}$ as inputs. 
    
    For a time-based TP-GMM, the cost is defined as root mean square error. For $\mu$ number of initial demonstrations, we represent the cost as:
    \begin{equation}\label{eq:rms}
        \mathcal{J}_{\text{RMS}} = \frac{1}{\mu}\sum^\mu_{i = 1}\sqrt{||\vy_i - \vxi_i||},
    \end{equation}  
    where $\vy_i$ is the reproduction of initial demonstrations from the TP-GMM and $\vxi_i$ is the initial expert demonstration, both in the $i$\textsuperscript{th} instance.
    
    For trajectory-based TP-GMM, we define the cost of the TP-GMM as the normalized distance computed by dynamic time warping (DTW) between the reproduction and the expert demonstrations.
    
    Give the reproduction $\vy \in \mathbb{R}^N$ and corresponding expert demonstration $\vxi_d \in \mathbb{R}^M$ (note for trajectory-based TP-GMM, $N$ does not necessarily equals $M$), DTW tries to find a warping function $\phi(K)$, where $K = 1,2,\ldots, T$:
    \[
    \phi(K) = \big( \phi_{\vy}(K), \phi_{\vxi}(K) \big),
    \] 
    where $\phi_{\vy}(K) \in \vy$, and $\phi_{\vxi}(K) \in \vxi$. The warping function maps the data in $\vy$ to the data in $\vxi$ with minimum dissimilarity $d$. The dissimilarity $d$ between two data points is defined as their Euclidean distance.
    The cost function for DTW is defined as the average accumulated dissimilarity between warped $\vy$ and $\vxi$:
    \begin{equation}\label{eq:dissimilarity}
        d_\phi (\vy, \vxi) = \sum_{K = 1}^T d\big(\phi_{\vy}(K), \phi_{\vxi}(K)\big) \zeta_\phi(K) / Z,
    \end{equation}
    where $\zeta_\phi(K)$ is a weighting coefficient and $Z$ is the normalization constant (we refer readers to \cite{giorgino2009computing} for a detailed definition).
    The warping function is computed by minimizing the cost defined in \eqref{eq:dissimilarity}:
    \begin{equation}
        D(\vy, \vxi) = \min_{\phi} d_\phi (\vy, \vxi)
    \end{equation}
    
    For $\mu$ number of initial demonstrations (each with a distinctive situation), we represent the cost for trajectory-based TP-GMM as:
    \begin{equation}\label{eq:cost}
        \mathcal{J}_{\text{DTW}} = \frac{1}{\mu}\sum^\mu_{i = 1} D\big(\vy(i),~\vxi_d(i)\big)
    \end{equation}
    We choose the cost depending on the type of TP-GMM and use it as a selection criterion for deciding whether we should update the dataset and TP-GMM or not.   
    
\begin{figure}[t]
        \centering
        \subfloat[Expert]{\includegraphics[width=0.24\columnwidth]{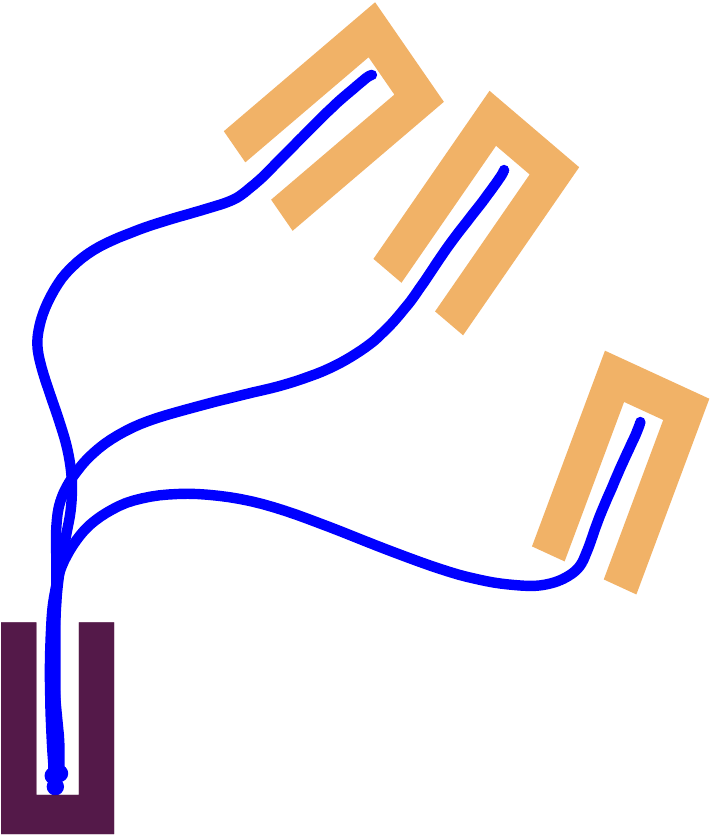}\label{fig:expert_demostrations}}
        \hfill
        \subfloat[Noise]{\includegraphics[width=0.24\columnwidth]{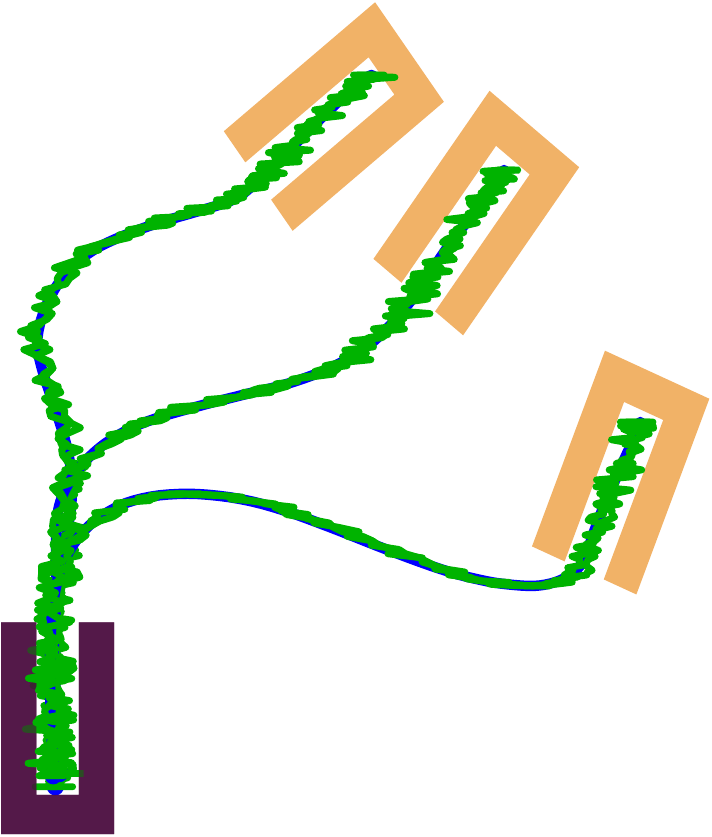}\label{fig:expert_demonstration_noise}} \hfill
        \subfloat[RF]{\includegraphics[width=0.24\columnwidth]{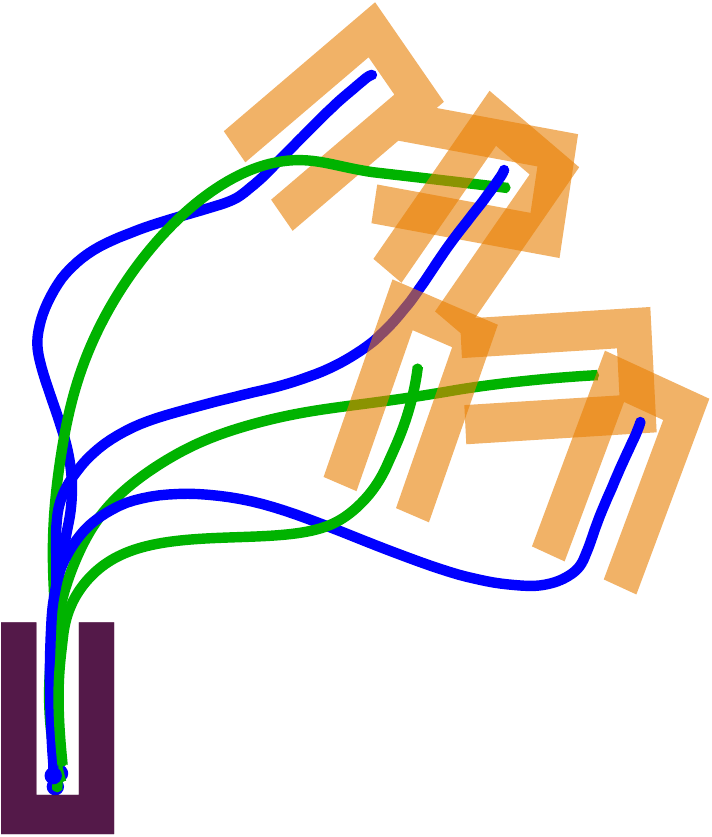}\label{fig:expert_demonstration_new_sit}} \hfill
        \subfloat[RF+noise]{\includegraphics[width=0.24\columnwidth]{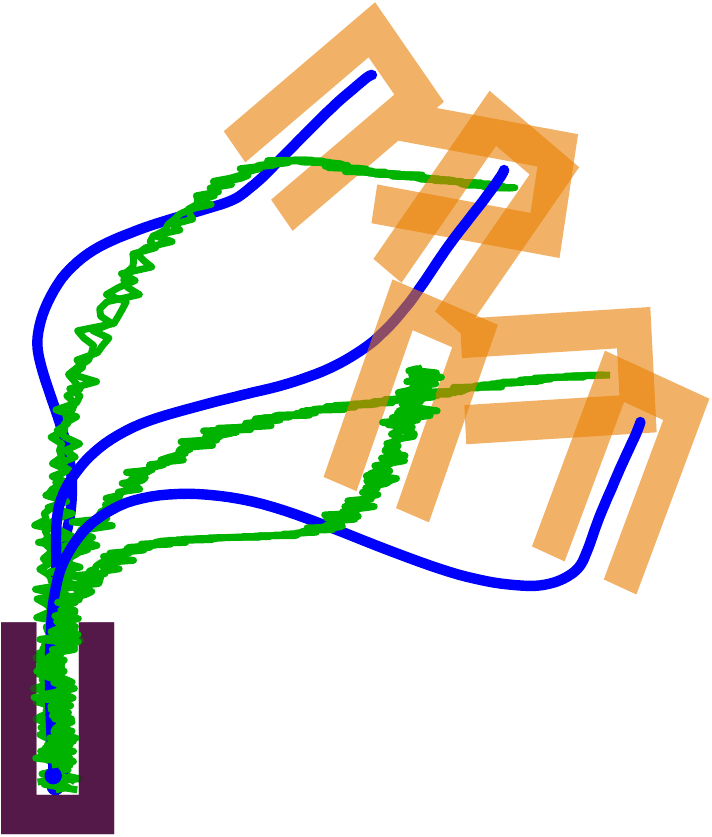}\label{fig:expert_demonstration_new_sit_noise}}
        \caption{Simulated movement tasks with expert demonstrations in blue and synthetic movement data in green: (a) Task representation and original expert demonstration, (b) original demonstrations with added noise, (c) generated motions in new situations, (d) generated noisy motions in new situations.}
    \end{figure}

    \textbf{\emph{Generate Synthetic Data:}} In this step, we generate new motion data to augment original demonstration for policy improvement. We present $3$ different data generation methods for Alg. \ref{alg} and introduce each subsequently:
        \begin{itemize}
        \item \emph{Noise:} Injecting white noise to the original expert demonstrations (Fig.~\ref{fig:expert_demonstration_noise}), as in \cite{zhao2021augmenting},
        \item \emph{RF:} TP-GMM generated data in random new situations (Fig.~\ref{fig:expert_demonstration_new_sit}),
        \item \emph{RF+noise:} TP-GMM generated data in random new situations adding a white noise (Fig.~\ref{fig:expert_demonstration_new_sit_noise}), 
    \end{itemize}
    
    For \emph{Noise}, we consider injecting white noise into the expert demonstration for generating the synthetic data. Noise injection on the training data is a known method for improving learning outcomes \cite{grandvalet1997noise}. It has been applied on Deep Neural Networks (DNNs) for achieving better generalization capability \cite{matsuoka1992noise} and robustness \cite{he2019parametric}. 
    
    \emph{RF} generates random reference frames that satisfies task constraints such as kinematic limits to create a new situation. Then applies the TP-GMM to the new situation for generation of new demonstration data.
    
    In TP-GMM, reference frames are attached to movable entities that are relevant for the task. Depending on the task space and the property of the entity, the frame orientation and translation are bounded for the task. For instance, orientation of the frames can be expressed in Euler angles with maximum and minimum values:
    \begin{equation}\label{eq:orientation_limits}
        \alpha \in [\alpha_{\min},\alpha_{\max}],~\beta \in [\beta_{\min},\beta_{\max}],~\gamma \in [\gamma_{\min},\gamma_{\max}],
    \end{equation}
    whereas translation limitation are expressed with:
    \begin{equation}\label{eq:translation_limits}
        \vb \in [\vb_{\min}, \vb_{\max}].
    \end{equation}
    
    We perform uniform random sampling between the limits in \eqref{eq:orientation_limits} and \eqref{eq:translation_limits} to generate reference frames that satisfy kinematic limits of the task. Once we have the new situation, a new motion data $\mathcal{D}_n$ can be obtained from the TP-GMM. 
    
    In the \emph{RF}, the new training data is generated by the fusion of different Gaussian distributions with new task parameters. It can not be generated by the individual GMM in each frame of the old TP-GMM. Adding the data to the training dataset helps to better understand the distributions in each frame, rather than reinforcing the original policy.
    
    \emph{RF+noise} is the combination of \emph{Noise} and \emph{RF}. It injects white noise into the data generated from \emph{RF} and use it for augmenting the original training data.

    We then augment the training dataset with the synthetic data, and retrain a TP-GMM based on the new dataset. Afterwards, the cost $\mathcal{J}$ of the new TP-GMM is compared with the old one. If the cost is reduced, we update the dataset and the TP-GMM. If not, we keep the old dataset and TP-GMM and go back to synthetic data generation step. Note that the cost is only calculated based on the expert demonstrations, and does not include the algorithm generated demonstrations. 
    
    \textbf{\emph{Outputs:}} The algorithm has two termination conditions. The maximum number of the demonstrations in the training dataset $\mathcal{D}$ denoted by $M$ ($M > \mu$) and the number of maximum iterations $L$. If the algorithm reaches the maximum iteration but fails to add any new demonstrations, one could either set a larger $L$, or provide additional demonstrations. In this way, the number of expert demonstrations required can be decided interactively.
    The outputs is the final TP-GMM that has the lowest cost after the dataset augmentation.

\section{Simulation and Analysis}\label{sec:simulation}
\begin{figure}[!h]
        \centering
        \subfloat[]{\includegraphics[width=0.22\columnwidth]{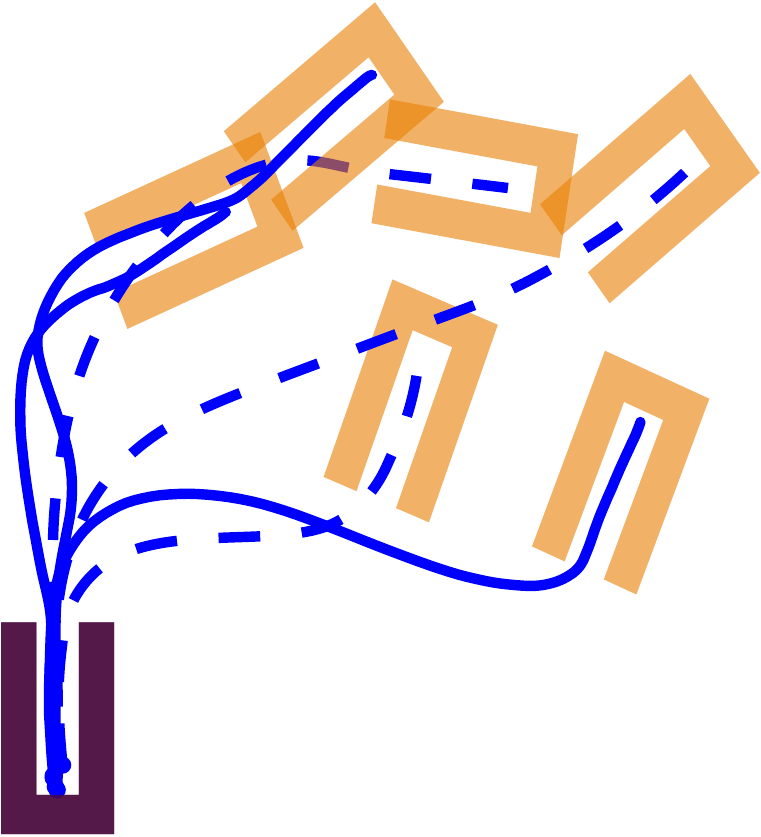}\label{fig:training_validation}} 
        \hfill
        \subfloat[]{\includegraphics[width=0.18\columnwidth]{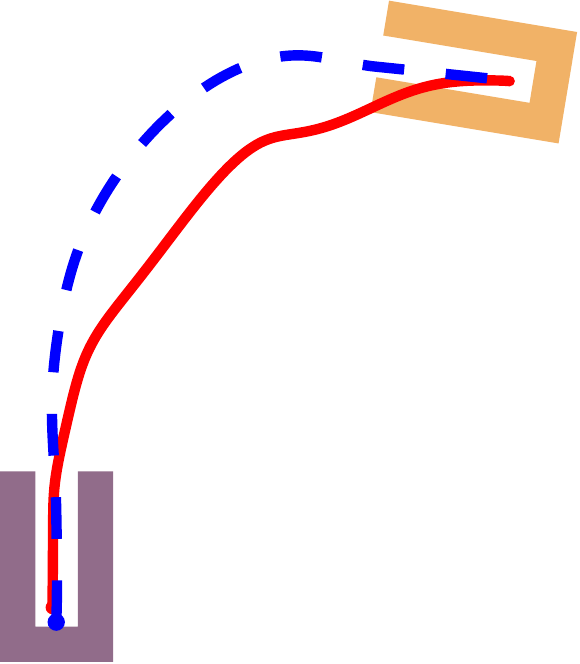}} 
        \hfill
        \subfloat[]{\includegraphics[width=0.15\columnwidth]{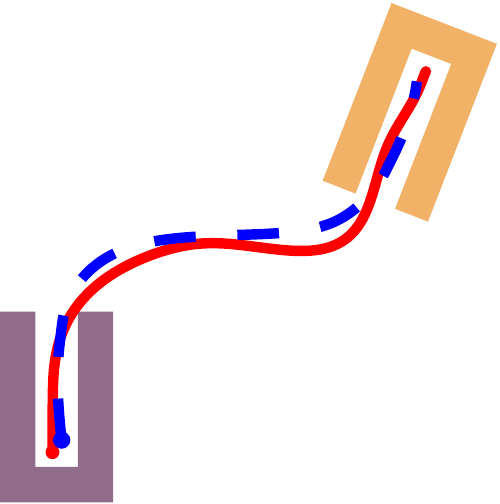}} 
        \hfill
        \subfloat[]{\includegraphics[width=0.22\columnwidth]{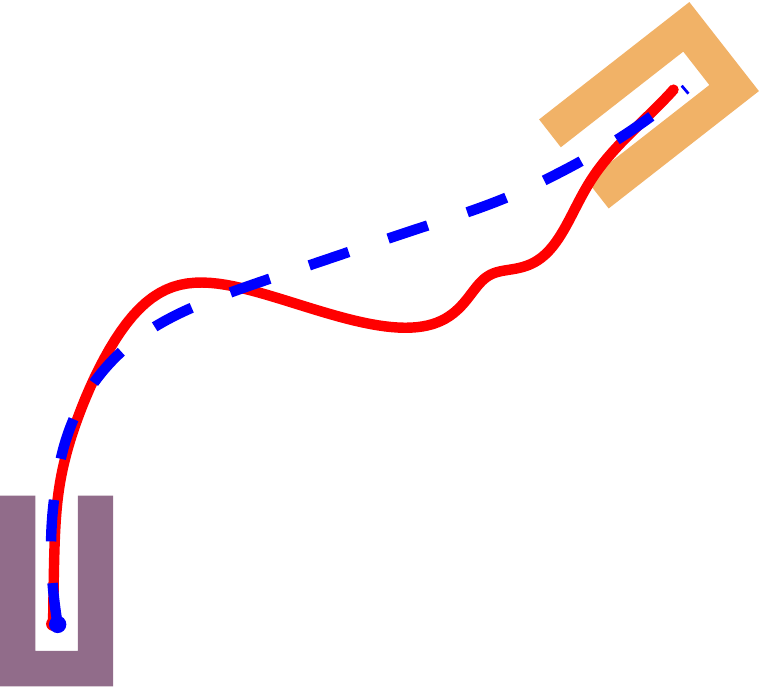}}
        \caption{The training (blue) and validation (dashed blue) set in one image in (a). Figure (b) -- (d) show the policy generated with initial TP-GMM (in red) compared with expert demonstrations (in dashed blue) in the validation set.}
        \label{fig:init_generalization}
    \end{figure}

    In this section, we analyze data generation methods proposed in Sect. \ref{sec:methods} for Alg. \ref{alg} in terms of cost reduction and policy generalization on a simulated task\footnote{The simulation results in this section is coded based on pbdlib: \url{https://gitlab.idiap.ch/rli/pbdlib-matlab/}}. The task considers a 2D movement from the start (the grey U-shape box) to the goal (the yellow U-shape box) as shown in Fig. \ref{fig:expert_demostrations}. The shape imposes constraints for the motion. We use in total six expert demonstrations, and split them into a training and validation set (Fig. \ref{fig:training_validation}). Each contains three expert demonstrations with different positions and orientations of the target. Subsequently, a time-based TP-GMM can be learned from the training set.
  
\begin{figure}[t]
        \centering
        \subfloat[Initial]{\includegraphics[width=0.24\columnwidth]{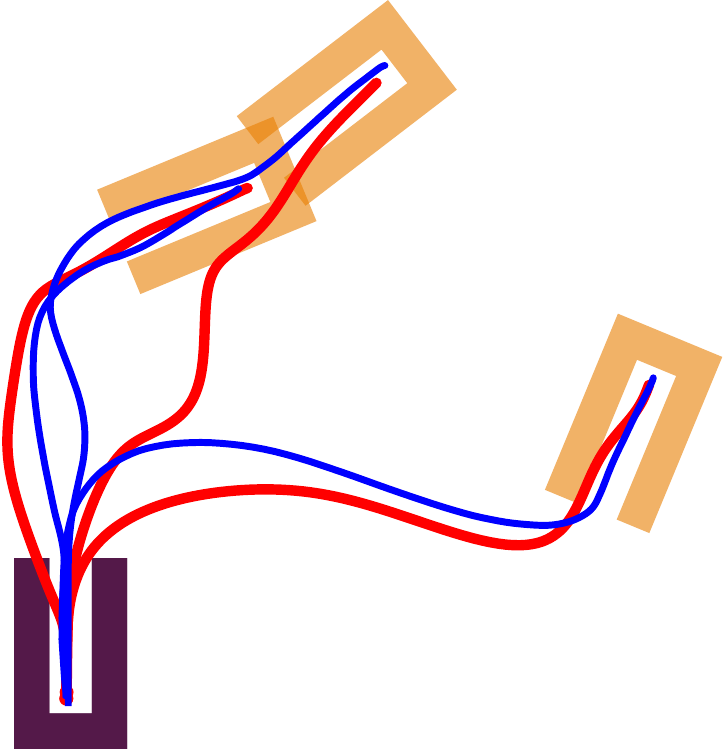}\label{fig:repo_original}}
        \hfill
        \subfloat[Noise]{\includegraphics[width=0.24\columnwidth]{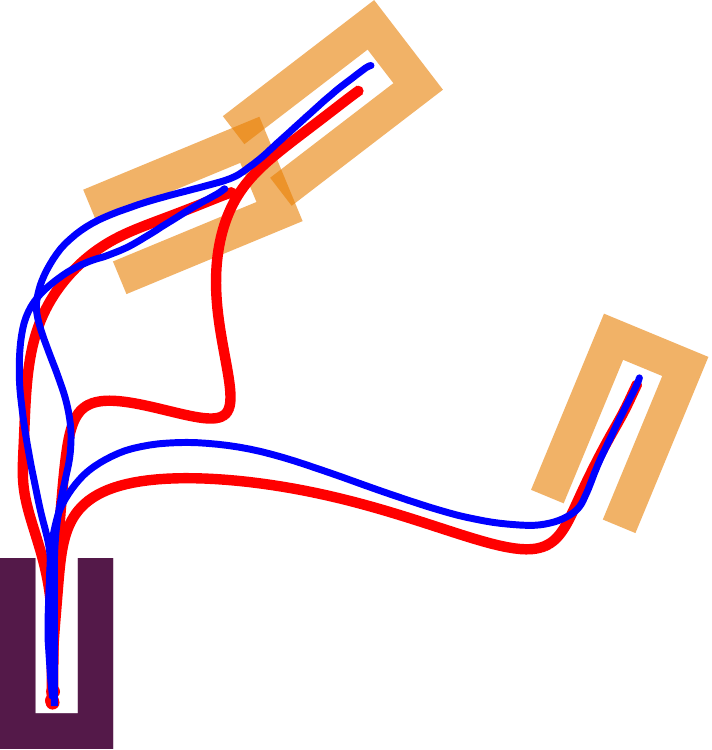}\label{fig:repo_noisy}} 
        \hfill
        \subfloat[RF]{\includegraphics[width=0.24\columnwidth]{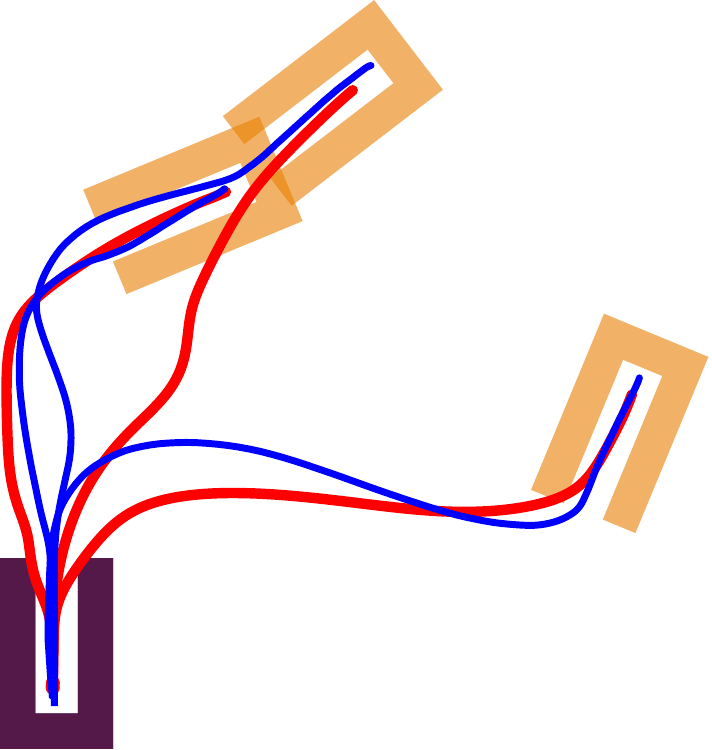}\label{fig:repo_new_sit}} 
        \hfill
        \subfloat[RF+noise]{\includegraphics[width=0.25\columnwidth]{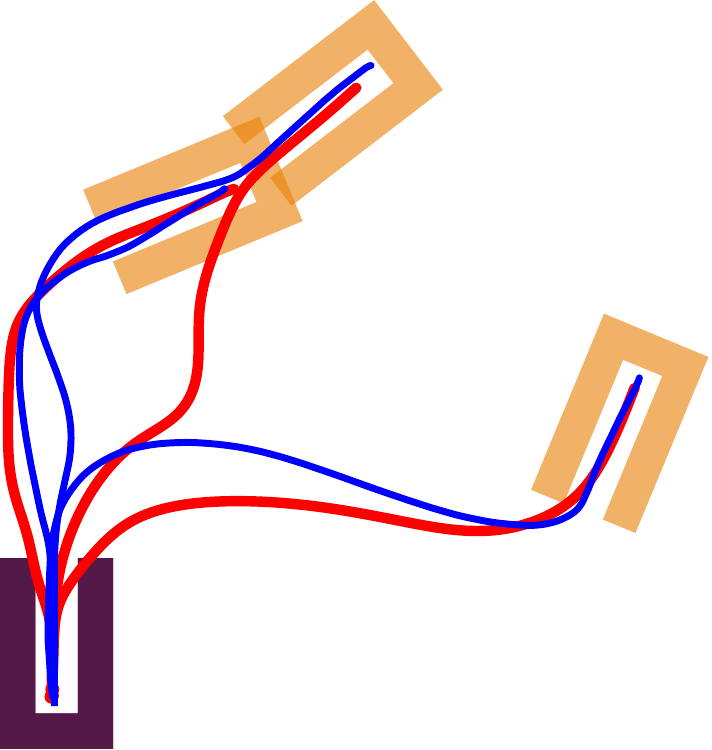}\label{fig:repo_noisy_new_sit}} 
        \caption{Reproduction of expert demonstrations in training set with original selection, the expert demonstrations are in blue and reproduction in red.}
        \label{fig:repo}
    \end{figure}

    We train the initial TP-GMM using the training set (shown in Fig.~\ref{fig:training_validation} in blue lines) with $8$ Gaussians. The reproduction of the initial TP-GMM on the training set is presented in Fig. \ref{fig:repo_original}. We define a separate validation set shown in dashed blue on Fig.~\ref{fig:training_validation}.
    
    Fig. \ref{fig:init_generalization} demonstrates the generalization capability of the initial TP-GMM by presenting its policy generation on each new situation in the validation set. We can observe that the generalization to new situations are not satisfactory for the initial TP-GMM.
    
    \begin{table*}[!h]
        \caption{Training and validation cost for the initial and improved TP-GMM}
         \centering
         \begin{tabular}{c|c|c|c|c|c|c|c}
         \hline
         \multirow{2}{4em}{} & \multicolumn{3}{c|}{Original Selection} & \multicolumn{3}{c|}{Generalization Selection} & \multirow{2}{5em}{Init. TP-GMM} \\ \cline{2-7}
          & \emph{Noise} & \emph{RF} & \emph{RF+Noise} & \emph{Noise} & \emph{RF} & \emph{RF+Noise} &  \\
          \hline
          training cost & \cyan{1.97 (91\%)} & \cyan{2.08 (96\%)} & \cyan{2.04 (94\%)} & \red{2.21 (102\%)} & \red{2.41 (111\%)} & 2.17 (100\%) & 2.17 (100\%) \\
          \hline
           validation cost & \red{2.50 (118\%)} & \cyan{1.85 (88\%)} & \cyan{1.84 (87\%)} & \cyan{1.48 (70\%)} & \cyan{1.79 (85\%)} & \cyan{1.82 (86\%)} & 2.11 (100\%) \\
          \hline
         \end{tabular}
         \label{tab:alg_analysis}
     \end{table*}
     
    We evaluate Alg.~\ref{alg} on the initial TP-GMM with two selection criteria, the two selection criteria are differed in their cost computation:
    \begin{itemize}
        \item Original selection (criterion): This criterion is what originally proposed in the algorithm where the cost is computed with regard to the training set.
        \item Generalization selection (criterion): This criterion considers a cost computed with regard to the validation set which contains three new expert demonstrations in new situations\footnote{Note that this selection criterion is only used as an ablation for the best-case performance in the new situations. In practice it requires more demonstrations than the original selection criterion.}. 
    \end{itemize}

    For each selection (criterion), we improve the initial TP-GMM models using three methods: namely \emph{Noise}, \emph{RF} and \emph{RF+Noise}. The termination condition for the algorithm is set to be $M = 8$ (max number of demonstrations) and $L = 50$ (max iterations). For the noise injection, the signal to noise ratio (SNR) is set to 30 decibels.  We run the algorithm with each data generation methods $20$ times and save the models with the maximum cost reduction. 
    
    We hence get $6$ improved TP-GMM models. We then compute the cost that defined in \eqref{eq:rms} for the training and validation set for each improved model. While the training cost indicates the performance of policy in reproducing the expert demonstrations, the validation cost measures the policy generalization in new situations. The results are summarized in Tab.~\ref{tab:alg_analysis}. 
    We use the cost of the initial TP-GMM as the reference quantity and provides a percentage value besides each cost value. The increased cost is marked with red and decreased with cyan in the table.
    
    
    From Tab. \ref{tab:alg_analysis} we observe that in the original selection, all three data augmentation methods can reduce the cost on the training set. As the cost is reduced, they all show better performance in terms of reproduction of the training set as also shown in Fig.~\ref{fig:repo}. As for the generalization selection, since the criterion for selecting synthetic data is defined with regard to the validation set, the cost on training set remain close to the reference in \emph{Noise} and \emph{RF+Noise}, and is even higher in the case of \emph{RF}. For the increase cost, we suspect that adding more trajectories with the generalization selection is likely to make the final model over-fit to the validation set, thus the cost on the training set increases. However, due to the page limit, we cannot discuss this in detail here.
    
\begin{figure}[!h]
        \centering
        \subfloat[]{\includegraphics[width=0.24\columnwidth]{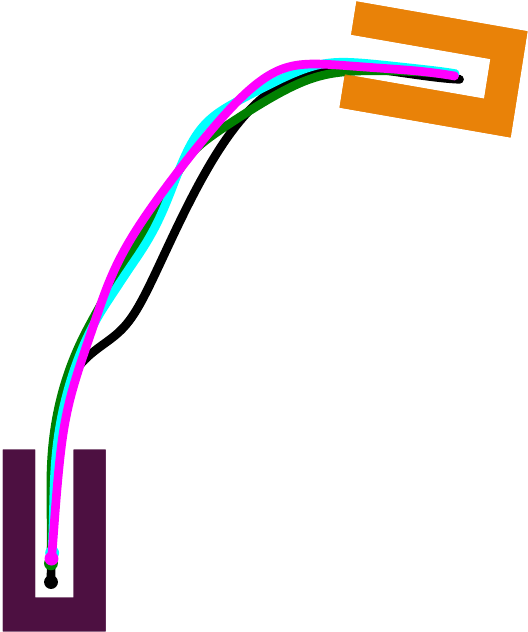}} 
        \hfill
        \subfloat[]{\includegraphics[width=0.2\columnwidth]{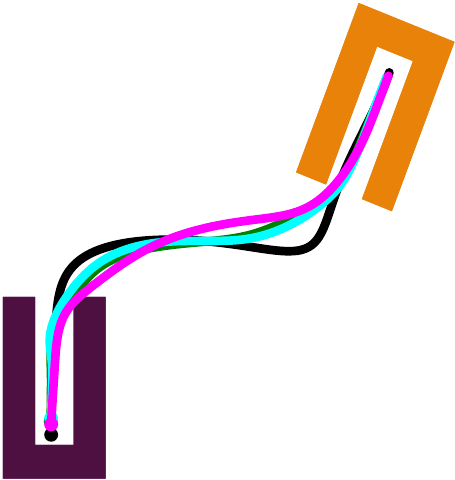}} 
        \hfill
        \subfloat[]{\includegraphics[width=0.3\columnwidth]{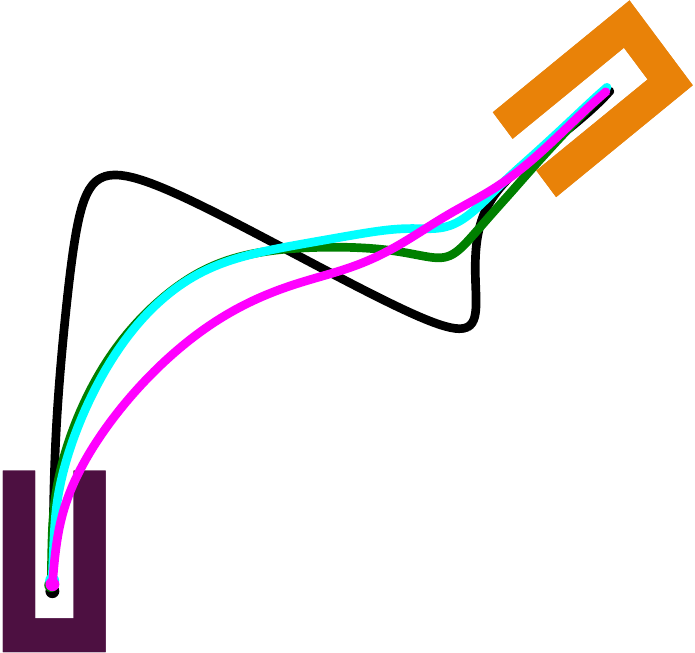}\label{fig:generalization_overfit}}
        \caption{Generalization to validation set by TP-GMM produced with \emph{Noise} (black), TP-GMM produced with \emph{RF} (green),  TP-GMM produced with \emph{RF+noise} (cyan), all using the original selection criterion and TP-GMM produced with \emph{RF+Noise} using the generalization selection criterion (magenta).}
        \label{fig:generalization}
    \end{figure}

    With the original selection, \emph{Noise} is able to reduce the cost the most. It is reasonable as the augmented data is the demonstrations from the training set plus white noise. Reduction of the training cost is one indicator of the performance. More importantly, generalization to new situations is the core capability of TP-GMM.
    In that regard, \emph{Noise} performed poorly on the validation set with an increased cost, indicating overfitting to the training set. The phenomena can be visually confirmed in Fig.~\ref{fig:generalization_overfit} (black trajectories) where \emph{Noise} turns out to magnify some unwanted movements in the produced motions in the new situation.
    
    \emph{RF} and \emph{RF+noise}, on the other hand, provide a reasonable cost reduction on both training and validation sets. Such reduction is manifested in the improved motions in Fig. \ref{fig:generalization} as compared to the initial TP-GMM in Fig. \ref{fig:init_generalization}(b) -- (d). Although the cost reduction on the validation set is slightly higher than in the generalization selection, but they are comparable. This point is further illustrated in Fig.~\ref{fig:generalization} which shows that the motions produced from \emph{RF} and \emph{RF+noise} in the original selection is very similar to \emph{RF+Noise} in the generalization selection.

    Another way of cost reduction without generating new data is to increase the number of Gaussian $K$ in \eqref{eq:tp_gmm_model} used in the TP-GMM training phase, however, when initial expert demonstrations are few and sparse, the TP-GMM is likely to over-fit to the expert demonstrations, and has the possibility to magnify unwanted movements in the demonstration similar to \emph{Noise}.
    One good indicator for checking the fitting is the reconstruction error of demonstrations we defined in \eqref{eq:rms} and \eqref{eq:cost}. For selecting a reasonable $K$, as suggested in \cite{calinon2016tutorial}, methods such as Bayesian information criterion \cite{schwarz1978estimating} and Dirichlet process \cite{chatzis2012nonparametric} can be used. 
    
    In a nutshell, by explicitly using reproduction error as the selection criterion, all three data generation methods are able to generate synthetic motions for data augmentation to improve TP-GMM. While TP-GMM produced from \emph{Noise} tends to over-fit to original demonstrations and may magnify unwanted movements, the TP-GMM produced from \emph{RF} and \emph{RF+Noise} are improved over the original TP-GMM in reproduction and also generalization. The generalization is comparable with a generalization selection where the cost is computed with regard to $3$ new expert demonstrations in new situations.

\section{Robotic Experiments}\label{sec:result}
    We consider the task of dressing a short sleeve shirt onto one arm of a mannequin. We assume the arm posture is static during the dressing and the hand is already inside the armscye. The robot grasped on the shoulder area of the cloth. The dressing starts above the wrist and ends above the shoulder.
    
    The dressing assistance is a primary task that occurs everyday in elderly care, and has been considered in previous assistive robot research. Recently the authors of \cite{liprovably} address the problem from safety perspective and proposed a motion planning strategy that theoretically guarantees safety under the uncertainty in human dynamic models. Zhang et al., uses a hybrid force/position control with simple planning for dressing \cite{zhang2019probabilistic}, while \cite{clegg2020learning} use deep reinforcement learning (DRL) to simultaneously train human and robot control policies as separate neural networks using physics simulations. Although DRL yields satisfactory dressing policies, applying DRL in a real world setting is very difficult, especially if the task involves a human. LfD on the other hand, allows programming the robot by non-experts, thus can facilitate dressing skill learning: i.e., the robot can be programmed by healthcare workers. 
    
    LfD has been employed to encode dressing policies in the previous research. The dressing policy for a single static posture is encode  by a DMP in \cite{joshi2019framework}. When considering multiple different static postures and the policy needs to adapt, task-parameterized approaches becomes more relevant.  Sensory information and motor commands are combined in \cite{pignat2017learning} as a joint distribution in a hidden semi-Markov model, and then coupled with a task-parameterized model to generalize to different situations. 
    In \cite{hoyos2016incremental}, the authors propose $3$ techniques to incrementally updates the TP-GMM when new demonstration data is available and tested with a dressing assistance scenario. All techniques in \cite{hoyos2016incremental} require new expert demonstrations and focus on how to update the TP-GMM while our method focuses on how to generate synthetic data for policy improvement for TP-GMM. Both \cite{pignat2017learning, hoyos2016incremental} require several expert demonstrations (either at the beginning or incrementally) for generalizing the dressing task. In this section, we demonstrate that using our framework, the robot can learn to dress with only two expert demonstrations.
    
\begin{figure}[t]
        \centering
	    \subfloat[Case A posture \& path]{\includegraphics[width = 0.24\columnwidth]{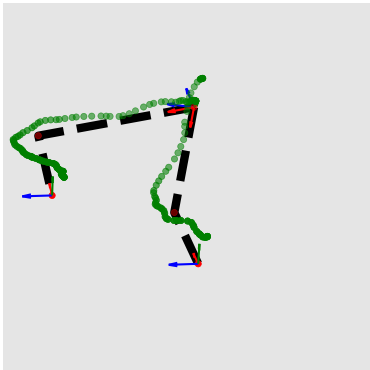}\label{fig:demo_close}}
	    \hfill
	    \subfloat[Case B posture \& path ]{\includegraphics[width = 0.24\columnwidth]{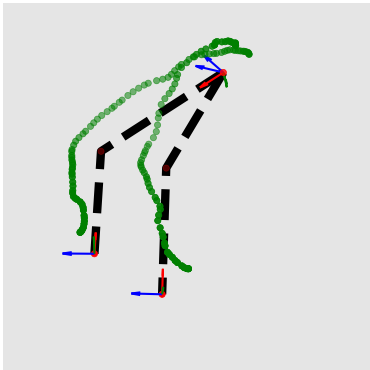}\label{fig:demo_far}}
	    \hfill
	    \subfloat[Case A arm postures]{\includegraphics[width = 0.24\columnwidth]{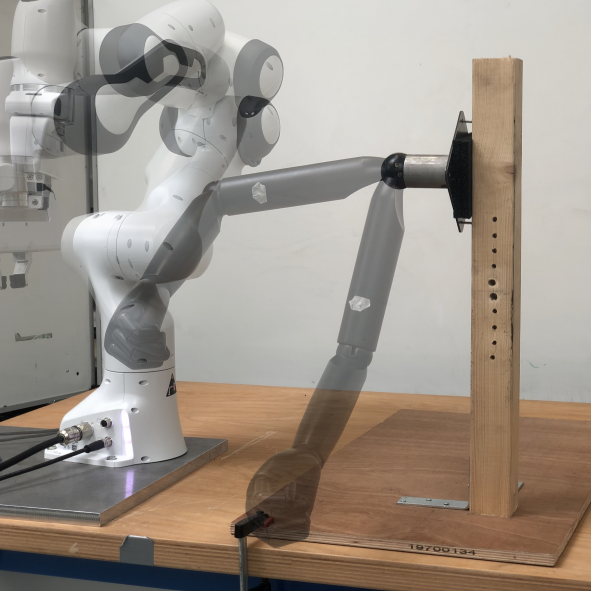}\label{fig:case_2}}
	    \hfill
	    \subfloat[Case B arm postures]{\includegraphics[width = 0.24\columnwidth]{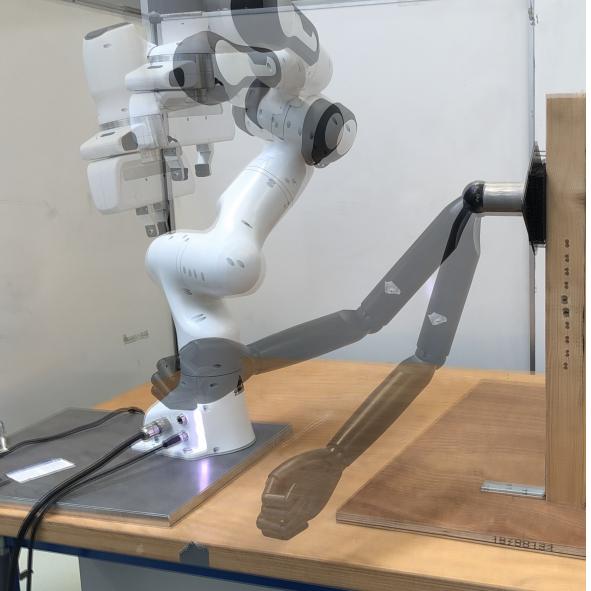}\label{fig:case_1}}
	    \caption{Two demonstrations dataset of the dressing task. Each dataset contains demonstrations in two different postures. Arm postures in (a) and (b) characterized by the dashed lines. Demonstrated path are scattered points in green. Local frames are depicted with XYZ (RGB arrows) coordinates at the shoulder and the wrist.}
        \label{fig:collect_demos}
    \end{figure}

    The dressing trajectory needs to adapt to different arm postures. The postures are described with positions of shoulder $\vp_\mathrm{sh}$, elbow $\vp_\mathrm{el}$ and wrist $\vp_\mathrm{wr}$ on a static base frame $s$ which located at the robot base. Two frames are needed to fully describe the posture of an arm. One located at the shoulder, and the other located at the wrist. These two frames constitute the task parameters for the TP-GMM:
    \begin{equation}
        \{\vA_\mathrm{sh}, \vb_\mathrm{sh}\},~\{\vA_\mathrm{wr}, \vb_\mathrm{wr}\},
    \end{equation}
    where $\vb_\mathrm{sh} = \vp_\mathrm{sh}$, $\vb_\mathrm{wr} = \vp_\mathrm{wr}$. For two orientations $\vA_\mathrm{sh}$ and $\vA_\mathrm{wr}$, the $x$ axis is defined parallel to vector $\vp_\mathrm{sh}\vp_\mathrm{el}$ and $\vp_\mathrm{wr}\vp_\mathrm{el}$ respectively. Fig.~\ref{fig:collect_demos} shows different postures and their respective reference frames at the shoulder and wrist with coordinate depicted with red, green and blue (RGB) arrows.
    
     During experiments, we record only positions of the end-effector (EE) and the wrist, elbow, shoulder positions during demonstration. The latter is used for calculating the shoulder and wrist reference frames. The robot motion in a new situations is generated by GMR produced from a TP-GMM upon the situation specific task parameters.
    
    Since the dressing motion is not explicitly depending on time, we employ the GMM consisting of position and displacement, reference frames can be augmented accordingly to transform both position and displacement components:
    \begin{equation}
        \hat{\vp} =\begin{bmatrix}
        \vp \\
        \delta\vp
        \end{bmatrix} \in \mathbb{R}^6,~
        \hat{\vA} = \begin{bmatrix}
        \vA & \vnull \\
        \vnull & \vA
        \end{bmatrix} \in \mathbb{R}^{6 \times 6},~
        \hat{\vb} =\begin{bmatrix}
        \vb \\
        \vnull
        \end{bmatrix} \in \mathbb{R}^6
    \end{equation}
    Following the steps described in Sect.~\ref{sec:model_training}, we can obtain a TP-GMM in the form of \eqref{eq:tp_gmm_model}. The dressing motion is generated by integrating the output of the GMR conditioned on the current position.
    
     We test our algorithm on two cases. In each case, the demonstration dataset contains human demonstrations in two different postures. Fig.~\ref{fig:collect_demos} presents the postures and corresponding demonstrations in both cases. 
     
\begin{figure}[t]
         \centering
         \includegraphics[width=0.4\textwidth]{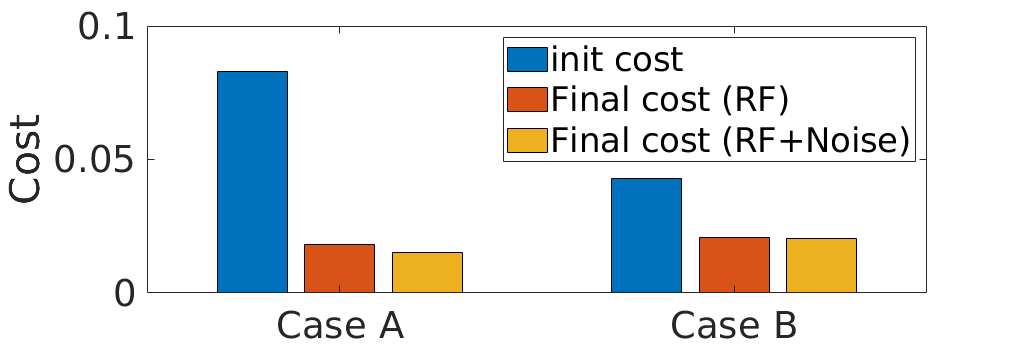}
         \caption{Cost reduction of Case A and B using proposed algorithm with \emph{RF} and \emph{RF+Noise} as the data generation method. Two cases have different initial cost as the demonstration data are different.}
         \label{fig:cost_reduction_experiments}
     \end{figure}

     The maximum number of demonstrations $M = 7$ and maximum iteration $L = 100$ are set for the algorithm. We train the TP-GMM with $4$ components. 
     Each initial demonstration dataset are tested on \emph{RF} and \emph{RF+noise} as the data generation methods. In \emph{RF+noise} the SNR is set to be 30 decibels. The cost reduction is indicate in Fig.~\ref{fig:cost_reduction_experiments}. 
     Two cases have different initial cost as the initial expert demonstrations are different. Both \emph{RF} and \emph{RF+noise} are able to reduce the cost. Similarly as in the simulation, both methods have compatible performance in terms of cost reduction. 
     In addition, we use a baseline TP-GMM trained with $7$ expert demonstrations for comparison.
     
     In addition, likewise in Sect. \ref{sec:simulation}, we test the generalization of the original and the final obtained TP-GMM. These TP-GMM are tested on $5$ different postures. We define the success condition as follows. If a trajectory is able to reach above and around shoulder and does not hit mannequin or stretch the cloth in the dressing process, we consider the dressing successful. The success rate (percentage of success out of total $5$ different postures) of the dressing policy generated from initial and final (synthetic data augmented) TP-GMM is summarized and compared with the baseline in Table \ref{tab:success_rate}. 
    \begin{table}[t]
        \caption{Success Rate of robot experiments}
         \centering
         \begin{tabular}{p{0.5cm}|p{2cm}|p{0.5cm}|p{2cm}|p{1cm}}
         \hline
         \multicolumn{2}{c|}{Case A} & \multicolumn{2}{c|}{Case B} & Baseline\\
         \hline
         init & final & init & final & final\\
         \hline 
         \multirow{2}{2em}{20\%} & 80\% (\emph{RF}) & \multirow{2}{2em}{20\%} & 60\% (\emph{RF}) & \multirow{2}{2em}{80\%} \\
          & 60\% (\emph{RF+noise}) & & 40\% (\emph{RF+noise}) & \\
         \hline
         \end{tabular}
         \label{tab:success_rate}
     \end{table}
     
     
     We observe that Case A improves more than Case B in terms of generalization. Take a closer look at the initial demonstrations of both cases, we observe that compared with Case A, in Case B, the initial two postures are more similar to each other and so are the resulting initial expert demonstrations. 
     This is also consistent with Fig. \ref{fig:cost_reduction_experiments} that the initial cost of Case B is less than Case A. The TP-GMM in Case B is probably over-fit to the data in  the region which might explains Case A outperforming B in terms of generalization. The best case scenario is able to match the baseline in terms of success rate. 
     
     The average run-time in experiment setup for our algorithm is around 20 seconds on a Linux-based i7-9750 PC, which is far less than the efforts and time of doing one extra expert demonstration\footnote{On average collecting one dressing demonstration on our current setup requires around $10$ seconds, without considering the time to change posture, and efforts in making the demonstration.}. 
     
     We notice even the baseline did not achieve 100\% in success rate. The failure case was stuck at the elbow. The reason might be that the TP-GMM we used only considers reference frames as task-parameter while the dressing path is dependent on additional latent variables. However, finding these variables is beyond the scope of this paper.    
     
\begin{figure}[t]
        \centering
	    \subfloat[Initial demonstration, shoulder frame, $xy$]{\includegraphics[width = 0.3\columnwidth]{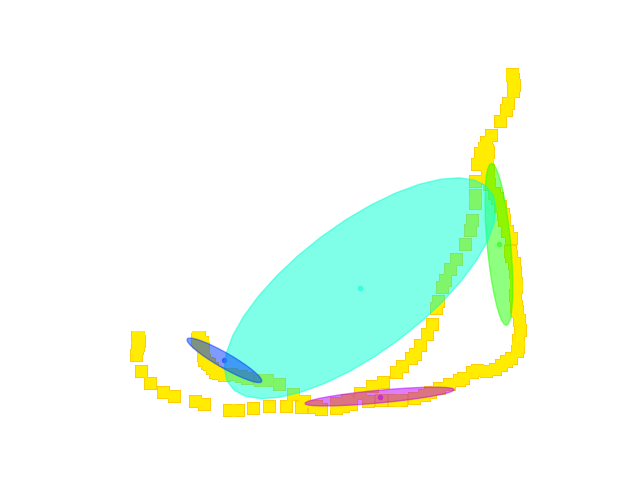}\label{fig:s_0_xy}}
	    \hfill
	    \subfloat[Init and $3$ synthetic data, shoulder frame, $xy$]{\includegraphics[width = 0.3\columnwidth]{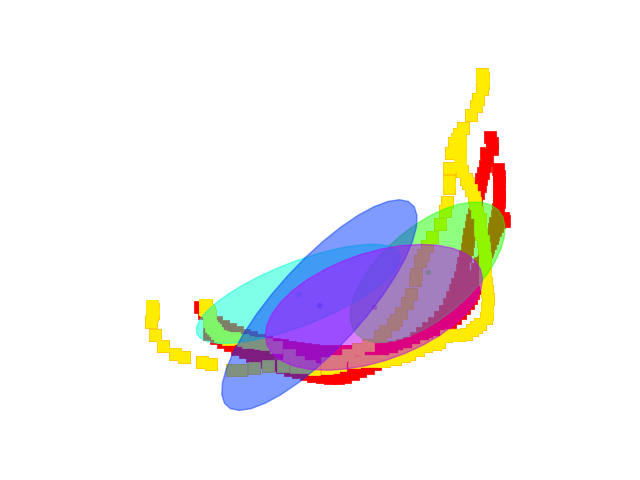}\label{fig:s_3_xy}}
	    \hfill
	    \subfloat[Init and $5$ synthetic data, shoulder frame, $xy$]{\includegraphics[width = 0.3\columnwidth]{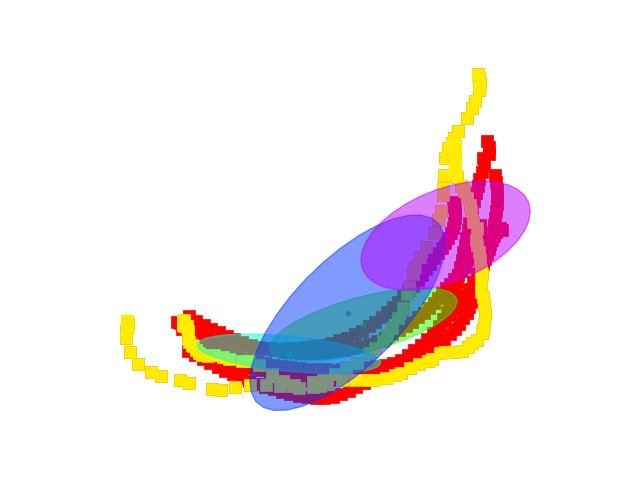}\label{fig:s_5_xy}} \\
	    \subfloat[Initial demonstration, shoulder frame, $xz$]{\includegraphics[width = 0.3\columnwidth]{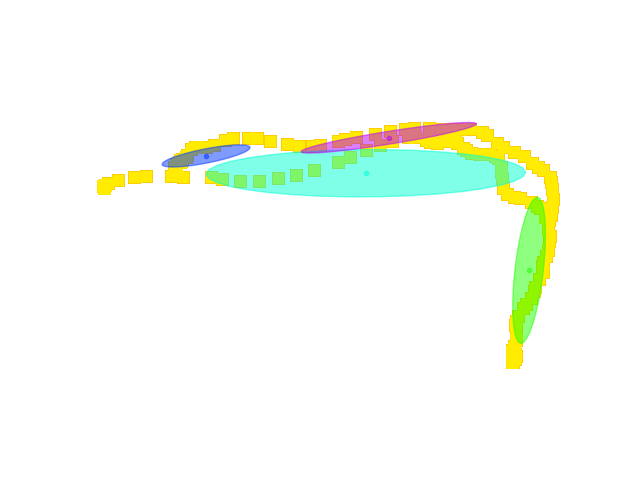}\label{fig:s_0_xz}}
	    \hfill
	    \subfloat[Init and $3$ synthetic data, shoulder frame, $xz$]{\includegraphics[width = 0.3\columnwidth]{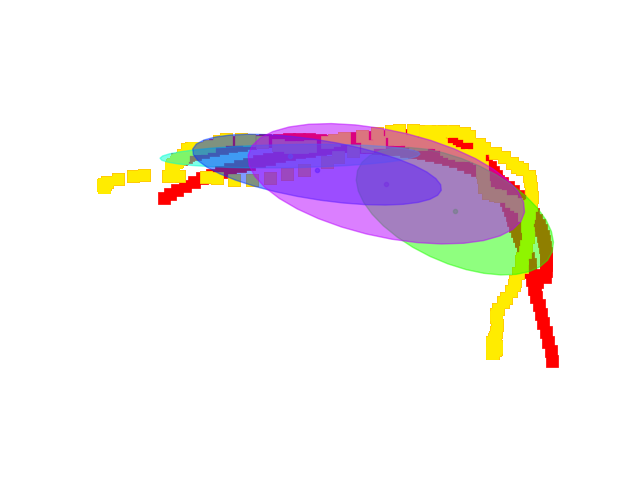}\label{fig:s_3_xz}}
	    \hfill
	    \subfloat[Init and $5$ synthetic data, shoulder frame, $xz$]{\includegraphics[width = 0.3\columnwidth]{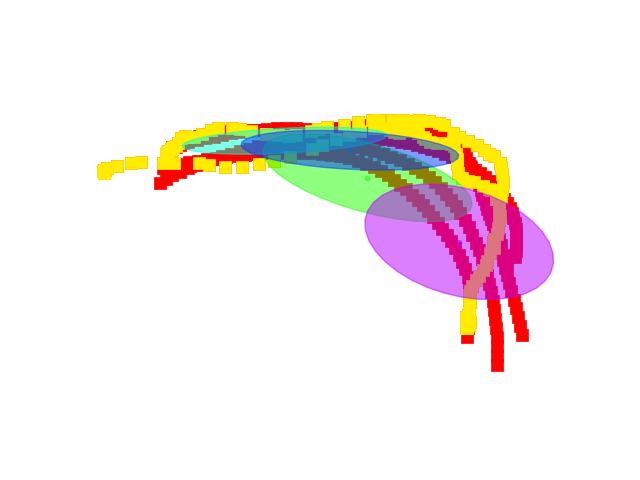}\label{fig:s_5_xz}} \\
	    \subfloat[Initial demonstration, wrist frame, $xy$]{\includegraphics[width = 0.3\columnwidth]{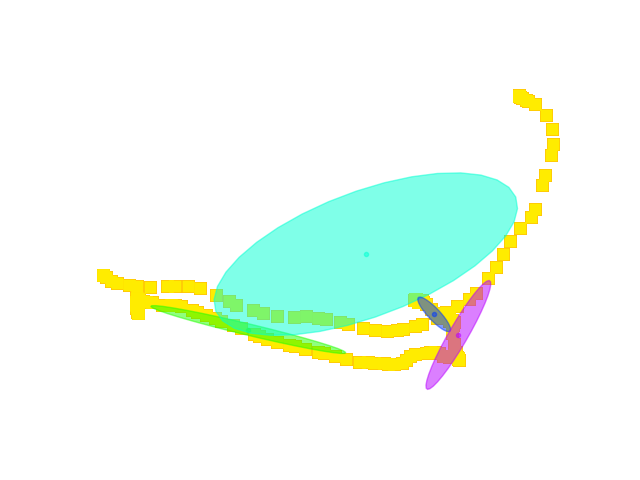}\label{fig:w_0_xy}}
	    \hfill
	    \subfloat[Init and $3$ synthetic data, wrist frame, $xy$]{\includegraphics[width = 0.3\columnwidth]{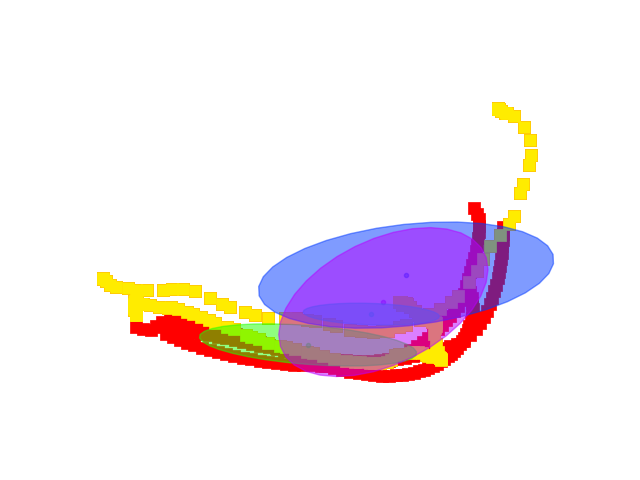}\label{fig:w_3_xy}}
	    \hfill
	    \subfloat[Init and $5$ synthetic data, wrist frame, $xy$]{\includegraphics[width = 0.3\columnwidth]{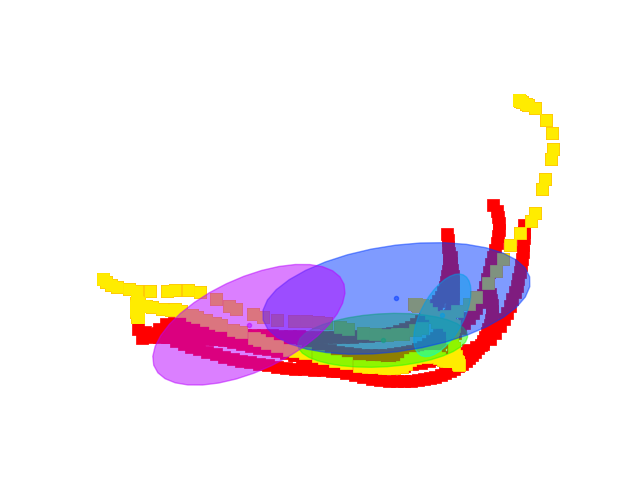}\label{fig:w_5_xy}} \\
	    \subfloat[Initial demonstration, wrist frame,$xz$]{\includegraphics[width = 0.3\columnwidth]{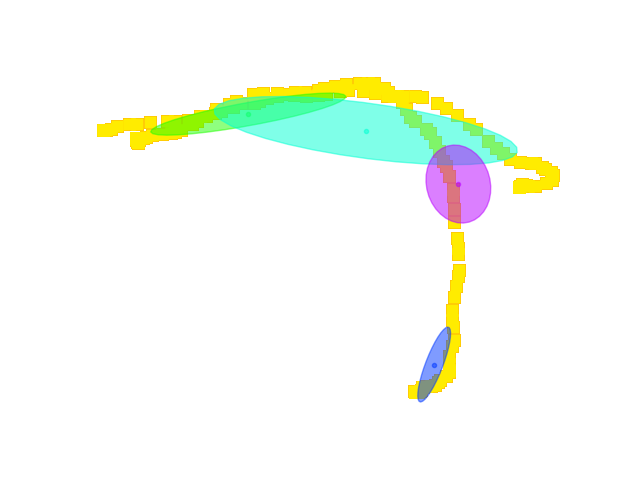}\label{fig:w_0_xz}} 
	    \hfill
	    \subfloat[Init and $3$ synthetic data, wrist frame, $xz$]{\includegraphics[width = 0.3\columnwidth]{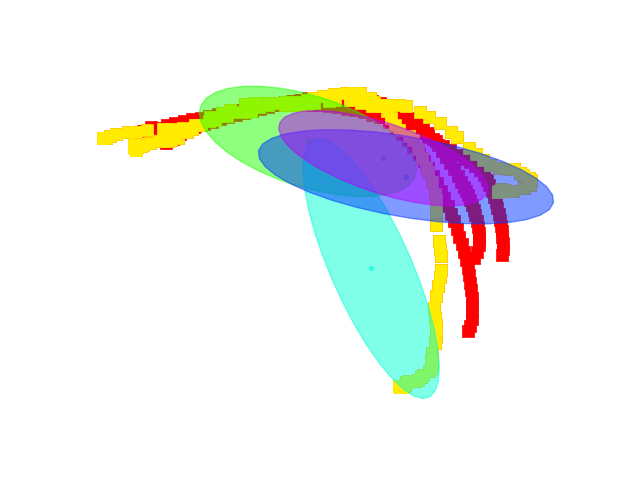}\label{fig:w_3_xz}} 
	    \hfill
	    \subfloat[Init and $5$ synthetic data, wrist frame, $xz$]{\includegraphics[width = 0.3\columnwidth]{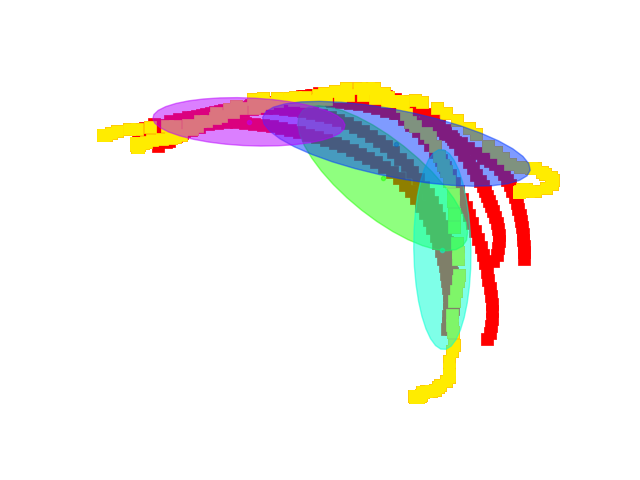}\label{fig:w_5_xz}} 
	    \caption{The synthetic demonstration data being added by algorithm and resulting changes of the GMM in each frame. The points in yellow and red are respectively original and synthetic demonstrations seen from 2D frames.}
        \label{fig:progress_data_adding}
    \end{figure}

     We use an example case with the best experiment performance (Case A with \emph{RF} as the data generation method) to show how added synthetic demonstrations change the distribution of data in each frames in terms of position data, and subsequently the mixture of Gaussian in 2D. We indicate correspondence with the same color in each column. Note that row-wise (which shows the changes of mixture models by adding synthetic data) the same color does not represents correspondence. Rows in Fig.~\ref{fig:progress_data_adding} show progressively new demonstration data being added by the algorithm and the resulting changes of the GMM in each frames. The data colored in yellow are initial expert demonstrations and red are synthetic demonstrations. The first two rows are $3D$ position data in the shoulder frame plotted with $xy$ and $xz$, the third and fourth are the data in the wrist frame. The ellipse represents the co-variance of each multivariate Gaussian.
    
     \begin{table*}[t]
        \centering
        \caption{Average cost reduction and number of trajectories generated}
        \begin{tabular}{p{5cm}|c|c|c|c|c|c}
            \hline
            No. of expert demo + synthetic demo (average discarded demo) & $2 + 0$& $2 + 1~(2.5)$ & $2 + 2~(21.2)$ & $2 + 3~(38.9)$ & $2 + 4~(47)$ & $2 + 5~(77)$  \\
            \hline
            Average Cost & $0.082$ & $0.068$ & $0.060$ & $0.051$ & $0.048$ & $0.046$ \\
            \hline
        \end{tabular}
        \label{tab:1}
    \end{table*}

    We run the the algorithm $20$ times to obtain the average cost and the number of synthetic trajectories needed for augmenting $n~(n = 1\ldots 5)$ number of new synthetic demonstrations. The results are presented in Table \ref{tab:1}. We can also observe that the reduction is larger when the first new demonstration is added, then becomes less significant when the number of demonstrations becomes more.
\section{Conclusion and Future Works}\label{sec:conclusion}
    Task-parameterized learning often requires creating multiple different situations for collecting demonstration thus increase the physical labour during data collection and the risk of having ambiguous demonstrations. We propose an algorithm for learning task-parameterized skills with a reduced number of demonstrations. The algorithm allows generation of new demonstrations that augment the original training dataset for improving the TP-GMM. We validate the algorithm in simulation and on real robot experiments with a dressing assistance task. 
    
    We observe that noise injection, creating new situations and the combination of both are possible data augmentation methods. Although noise injection provides the highest cost reduction, it has the possibility of magnifying unwanted movement when few demonstrations are available. The latter two perform similarly in terms of policy improvement while render a better generalization to new situations. In addition, distinctive initial demonstrations may contribute to better generalization performance of the algorithm. 
    
    TP-GMM needs a well-defined task space with reference frames which limits its application. For the dressing task, all TP-GMMs fail to dress at least one posture. That leads to a limitation of the current method: the exploration is done entirely in the synthetic domain which means the algorithm is not designed to handle environment feedback and can not explicitly use reward signals. As a consequence, the resulting TP-GMM will not able to resolve on the corner cases. RL-based policy search maybe promising in resolving corner cases as reported in ACNMP framework in \cite{akbulut2021acnmp} and adaptive ProMP in \cite{stark2019experience}. However, for dressing tasks, the reward needs to be carefully designed as reflected in \cite{clegg2018learning}. Alternatively, instead of discarding the trajectories, efforts could be made to exploit them \cite{akbulut2021reward} to further enhance the data efficiency.
    
    The current method considers a fix number of Gaussians. As the training data set is augmented with the algorithm, simultaneously increasing the number of Gaussian will probably result in a better approximation of data distribution in each frame and subsequently improve the policy even further.   


\bibliographystyle{IEEEtran}
\bibliography{refs}  

\end{document}